\documentclass[runningheads]{llncs}

\usepackage{eccv}

\usepackage{eccvabbrv}

\usepackage{graphicx}
\usepackage{booktabs}

\usepackage[accsupp]{axessibility}  

\usepackage{color}
\usepackage{xspace}
\usepackage{graphicx}
\usepackage{booktabs} 
\usepackage{multirow} 
\usepackage{utfsym}
\usepackage{url}
\usepackage{tcolorbox}

\usepackage{booktabs}
\usepackage{amsfonts}
\usepackage{nicefrac}
\usepackage{microtype}

\usepackage{color}
\usepackage{multirow}
\usepackage{multicol}
\usepackage{marvosym}
\usepackage{enumitem}
\usepackage{pifont}

\newcommand{\rebuttal}[1]{\textcolor{blue!90}{#1}}
\renewcommand{\rebuttal}[1]{#1}

\definecolor{citecolor}{HTML}{0071BC}
\definecolor{linkcolor}{HTML}{ED1C24}

\def\ours{GPT4RoI\xspace}
\definecolor{ForestGreen}{RGB}{34,139,34}
\definecolor{deemph}{gray}{0.6}

\usepackage{hyperref}

\usepackage{orcidlink}

\begin{document}

\title{GPT4RoI: Instruction Tuning Large Language Model on Region-of-Interest} 

\titlerunning{GPT4RoI}

\author{Shilong Zhang\inst{1}\orcidlink{0009-0005-4336-4941}\thanks{Equal contribution.} \and
Peize Sun\inst{1*}\orcidlink{0000-0003-0636-2224} \and
Shoufa Chen\inst{1*}\orcidlink{0000-0002-6126-2595} \and 
Min Xiao \inst{2}\orcidlink{} \\ 
  Wenqi Shao\inst{2}\orcidlink{} \and
  Wenwei Zhang\inst{2}\orcidlink{0000-0002-2748-4514} \and
     Yu Liu\inst{3}\orcidlink{} \and
       Kai Chen\inst{2}\orcidlink{} \and
          Ping Luo\inst{2}\orcidlink{0000-0002-6685-7950}
} 

\authorrunning{Shilong et al.}

\institute{ The University of Hong Kong \and Shanghai AI Laboratory  \and Alibaba Group}
\maketitle

\begin{abstract}
Visual instruction tuning large language model~(LLM) on image-text pairs has achieved general-purpose vision-language abilities. However, the lack of region-text pairs limits their advancements to fine-grained multimodal understanding. In this paper, we propose \textit{spatial instruction tuning}, which introduces the reference to the region-of-interest~(RoI) in the instruction. Before sending to LLM, the reference is replaced by RoI features and interleaved with language embeddings as a sequence. Our model \ours, trained on 7 region-text pair datasets, brings an unprecedented interactive and conversational experience compared to previous image-level models. (1) \textit{Interaction beyond language}: Users can interact with our model by both language and drawing bounding boxes to flexibly adjust the referring granularity. (2) \textit{Versatile multimodal abilities}: A variety of attribute information within each RoI can be mined by \ours, \textit{e.g.}, color, shape, material, action, \textit{etc}. Furthermore, it can reason about multiple RoIs based on common sense. On the Visual Commonsense Reasoning~(VCR) dataset, GPT4RoI achieves a remarkable accuracy of 81.6\%, surpassing all existing models by a significant margin (the second place is 75.6\%) and almost reaching human-level performance of 85.0\%. 
The code, dataset, and demo can be found at \url{https://github.com/jshilong/GPT4RoI}.
\end{abstract}
\section{Introduction}
\begin{figure}[ht]
  \centering
  \includegraphics[width=0.9\textwidth]{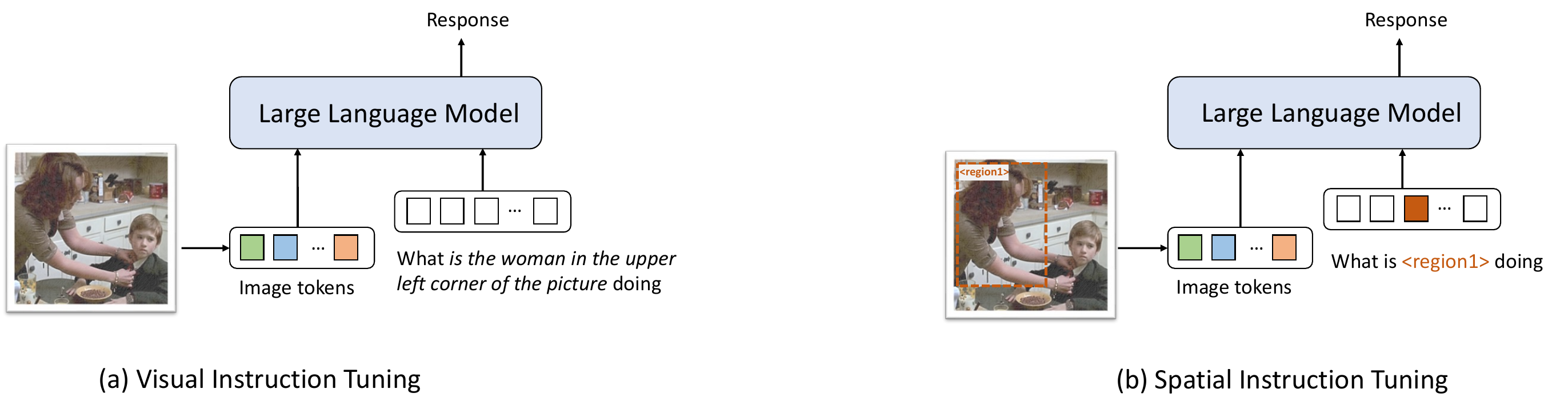}
  \caption{Comparison of visual instruction tuning on image-text pairs and spatial instruction tuning on region-text pairs. The bounding box and text description of each object are provided in region-text datasets. During training, the bounding box is from annotations, and in inference, it can be provided by the user or any off-the-shelf object detector.}
  \label{fig:teaser}
\vspace{-4mm}
\end{figure}

Recent advancements of large language models~(LLM) have shown incredible performance in solving natural language processing tasks in a human-like conversational manner, for example, commercial products ~\cite{openai2022chatgpt, anthropic2023claude, google2023bard, openai2023gpt4} and community open-source projects~\cite{touvron2023llama, touvron2023llamav2, alpaca, vicuna2023, du2022glm, sun2023moss}. Their unprecedented capabilities present a promising path toward general-purpose artificial intelligence models. Witnessing the power of LLM, the field of multimodal models~\cite{yang2023mm,huang2023language,girdhar2023imagebind,driess2023palm} is developing a new technology direction to leverage LLM as the universal interface to build general-purpose models, where the feature space of a specific task is tuned to be aligned with the feature space of pre-trained language models.

\begin{table}[t]
\setlength{\tabcolsep}{2.5pt}
\centering
\caption{Comparisons of vision-language models. Our \ours is an end-to-end model that supports region-level understanding and multi-round conversation.}
\begin{tabular}{lcccccc}
\toprule 
\multirow{2}{*}{Model} & \multirow{2}{*}{Image} & \multirow{2}{*}{Region} & \multirow{2}{*}{Multi-Region} & Multi-Round & End-to-End\\ 
& & & & Dialogue & Model 
\\ 
\hline
Visual ChatGPT~\cite{wu2023visualchatgpt} & 
\textcolor{ForestGreen}{\usym{2713}} & 
\textcolor{red}{\usym{2717}} &
\textcolor{red}{\usym{2717}} &
\textcolor{ForestGreen}{\usym{2713}} & 
\textcolor{red}{\usym{2717}} &
\\
MiniGPT-4~\cite{zhu2023minigpt} &
\textcolor{ForestGreen}{\usym{2713}} &
\textcolor{red}{\usym{2717}} &
\textcolor{red}{\usym{2717}} &
\textcolor{ForestGreen}{\usym{2713}} & \textcolor{ForestGreen}{\usym{2713}} &
\\
LLaVA~\cite{liu2023visual} &
\textcolor{ForestGreen}{\usym{2713}} & 
\textcolor{red}{\usym{2717}} &
\textcolor{red}{\usym{2717}} &
\textcolor{ForestGreen}{\usym{2713}} & \textcolor{ForestGreen}{\usym{2713}} &
\\
InstructBLIP~\cite{instructblip} &
\textcolor{ForestGreen}{\usym{2713}} & 
\textcolor{red}{\usym{2717}} &
\textcolor{red}{\usym{2717}} &
\textcolor{ForestGreen}{\usym{2713}} & \textcolor{ForestGreen}{\usym{2713}} &
\\
MM-REACT~\cite{yang2023mm} &
\textcolor{ForestGreen}{\usym{2713}} & 
\textcolor{ForestGreen}{\usym{2713}} &
\textcolor{ForestGreen}{\usym{2713}} &
\textcolor{ForestGreen}{\usym{2713}} & 
\textcolor{red}{\usym{2717}} &
\\
InternGPT~\cite{liu2023internchat} & 
\textcolor{ForestGreen}{\usym{2713}} & 
\textcolor{ForestGreen}{\usym{2713}} &
\textcolor{ForestGreen}{\usym{2713}} &
\textcolor{ForestGreen}{\usym{2713}} & 
\textcolor{red}{\usym{2717}} &
\\
VisionLLM~\cite{wang2023visionllm} & 
\textcolor{ForestGreen}{\usym{2713}} & 
\textcolor{ForestGreen}{\usym{2713}} & 
\textcolor{red}{\usym{2717}} &
\textcolor{red}{\usym{2717}} &
\textcolor{ForestGreen}{\usym{2713}} &
\\
CaptionAnything~\cite{wang2023caption} &
\textcolor{ForestGreen}{\usym{2713}} & 
\textcolor{red}{\usym{2717}} &
\textcolor{red}{\usym{2717}} &
\textcolor{red}{\usym{2717}} &
\textcolor{red}{\usym{2717}} &
\\
DetGPT~\cite{pi2023detgpt} &
\textcolor{ForestGreen}{\usym{2713}} & 
\textcolor{ForestGreen}{\usym{2713}} &
\textcolor{red}{\usym{2717}} &
\textcolor{ForestGreen}{\usym{2713}} & 
\textcolor{red}{\usym{2717}} &
\\ 
\midrule
\ours & 
\textcolor{ForestGreen}{\usym{2713}} &
\textcolor{ForestGreen}{\usym{2713}} & \textcolor{ForestGreen}{\usym{2713}} &
\textcolor{ForestGreen}{\usym{2713}} & \textcolor{ForestGreen}{\usym{2713}} &\\
\bottomrule  
\end{tabular}

\label{tab:c_vlm}

\end{table}
As one of the representative tasks, vision-and-language models align the vision encoder feature to LLM by instruction tuning on image-text pairs, such as MiniGPT-4~\cite{zhu2023minigpt}, LLaVA~\cite{liu2023visual}, InstructBLIP~\cite{instructblip}, etc. Although these works achieve amazing multimodal abilities, their alignments are only on image-text pairs~\cite{chen2015microsoft,sharma2018conceptual,changpinyo2021conceptual,ordonez2011im2text,schuhmann2021laion}, the lack of region-level alignment limits their advancements to more fine-grained understanding tasks such as region caption~\cite{krishna2017visual} and reasoning~\cite{zellers2019recognition}. To enable region-level understanding in vision-language models, some works attempt to leverage external vision models, for example, MM-REACT~\cite{yang2023mm}, InternGPT~\cite{liu2023internchat} and DetGPT~\cite{pi2023detgpt}, as shown in Table~\ref{tab:c_vlm}. However, their non-end-to-end architecture is a sub-optimal choice for general-purpose multi-modal models.

Considering the limitations of previous works, our objective is to construct an end-to-end vision-language model that supports a fine-grained understanding on region-of-interest. Since 
there is no operation that can refer to specific regions in current image-level vision-language models~\cite{zhu2023minigpt,liu2023visual,zhang2023llamaadapter,instructblip}, our key design is to incorporate references to bounding boxes into language instructions, thereby upgrading them to the format of \textit{spatial instructions}.  For example, as shown in Figure~\ref{fig:teaser}, when the question is \textit{``what is <region1> doing?''}, where the  \textit{<region1>} refers to a specific region-of-interest, the model will substitute the embedding of \textit{<region1>} with the region feature extracted by the corresponding bounding box. The region feature extractor can be flexibly implemented by RoIAlign~\cite{he2017mask} or Deformable attention~\cite{zhu2020deformable}.

To establish fine-grained alignment between vision and language, we involve region-text datasets in our training, where the bounding box and the text description of each region are provided. The datasets are consolidated from publicly available ones including COCO object detection~\cite{lin2014microsoft}, RefCOCO~\cite{yu2016modeling}, RefCOCO+~\cite{yu2016modeling}, RefCOCOg~\cite{mao2016generation}, Flickr30K entities~\cite{plummer2015flickr30k}, Visual Genome~(VG)~\cite{krishna2017visual} and Visual Commonsense Reasoning~(VCR)~\cite{zellers2019recognition}. These datasets are transformed into spatial instruction tuning format. Moreover, we incorporate the LLaVA150K dataset~\cite{liu2023visual} into our training process by utilizing an off-the-shelf detector to generate bounding boxes. This enhances our model's ability to engage in multi-round conversations and generate more human-like responses.

The collected datasets are categorized into two types based on the complexity of the text. First, the plain-text data contains object category and simple attribute information. It is used for pre-training the region feature extractor without impacting the LLM. Second, the complex-text data often contains complex concepts or requires common sense reasoning. We conduct end-to-end fine-tuning of the region feature extractor and LLM for these data.

Benefiting from spatial instruction tuning, our model brings a new interactive experience, where the user can express the question to the model with language and the reference to the region-of-interest. This leads to new capacities beyond image-level understanding, such as region caption and complex region reasoning.  As a generalist, our model GPT4RoI also shows its strong region understanding ability on popular benchmarks, including comprehensive image region understanding benchmark ViP-Bench~\cite{cai2023making}, the region caption task on Visual Genome~\cite{krishna2017visual}, the region reasoning task on Visual Commonsense Reasoning~\cite{zellers2019recognition} (VCR). Especially noteworthy is the performance on the most challenging VCR dataset, where GPT4RoI achieves an impressive accuracy of 81.6\%, 6 points ahead of the second-place and nearing the human-level performance benchmarked at 85.0\%.

In summary, our work makes the following contributions:
\begin{itemize}[leftmargin=*]
    \item We introduce spatial instruction, combining language and the reference to region-of-interest into an interleave sequence, enabling accurate region referring and enhancing user interaction.
    \item  By spatial instruction tuning LLM with region-text datasets, our model can follow user instructions to solve diverse region understanding tasks, such as region caption and reasoning. 
    \item Our method, as a generalist, outperforms the previous state-of-the-art approach on a wide range of region understanding benchmarks.
        
\end{itemize}

\section{Related Work}
\subsection{Large Language Model}

The field of natural language processing (NLP) has achieved significant development by the high-capability large language model (LLM). The potential of LLM is first demonstrated by pioneering works such as BERT~\cite{devlin2018bert} and GPT~\cite{radford2018gpt}. Then scaling up progress is started and leads to a series of excellent works, for example, T5~\cite{raffel2020t5}, GPT-3~\cite{brown2020gpt3}, Flan-T5~\cite{chung2022flant5}, PaLM~\cite{chowdhery2022palm}, etc. With the growth of training data and model parameters, this scaling-up progress brings a phenomenal product, ChatGPT~\cite{openai2022chatgpt}. By generative pre-trained LLM and instruction tuning~\cite{ouyang2022instructgpt} on human feedback, ChatGPT shows unprecedented performance on conversations with humans, reasoning and planning tasks~\cite{mu2023embodiedgpt,yang2023pave,bubeck2023sparks}, etc.

\subsection{Large Vision-Language Model}

To utilize high-performance LLM to build up vision-language models, LLM as task coordinator is proposed. Given the user instruction, LLM parses the instruction and calls various external vision models. Some representative works are Visual ChatGPT~\cite{wu2023visualchatgpt}, ViperGPT~\cite{suris2023vipergpt}, MM-REACT~\cite{yang2023mm}, InternGPT~\cite{liu2023internchat}, VideoChat~\cite{li2023videochat}, etc. Although these models largely expand the scope of multimodal models, they depend on external vision models and these non-end-to-end architectures are not the optimal choice for multi-modal models. To obtain end-to-end vision-language models, instruction tuning LLM on image-text pairs is proposed to align visual features with LLM and accomplish multimodal tasks in a unified way, for example, Flamingo~\cite{alayrac2022flamingo}, MiniGPT-4~\cite{zhu2023minigpt}, LLaVA~\cite{liu2023visual}, LLaMa-Adapter~\cite{zhang2023llamaadapter}, InstructBLIP~\cite{instructblip}, MM-GPT~\cite{gong2023multimodalgpt}, VPGTrans~\cite{zhang2023transfer}, etc. These models achieve amazing image-level multimodal abilities, while several benchmarks such as  LVLM-eHub~\cite{xu2023lvlmehub} and MMBench~\cite{liu2023mmbench} find that these models still have performance bottlenecks when need to be under specific region reference. Our \ours follows the research line of visual instruction tuning and moves forward region-level multimodal understanding tasks such as region caption~\cite{krishna2017visual} and reasoning~\cite{zellers2019recognition}.

\subsection{Region-Level Image Understanding}
For region-level understanding, it is a common practice in computer vision to identify potential regions of interest first and then do the understanding. Object detection~\cite{ren2015faster, carion2020end, zhu2020deformable, zang2023contextual} tackles the search for potential regions, which are generally accompanied by a simple classification task to understand the region's content.  To expand the object categories, ~\cite{kamath2021mdetr, liu2023grounding, zhou2022detecting, li2021grounded} learn from natural language and achieve amazing open-vocabulary object recognition performance. Region captioning~\cite{johnson2015densecap, CVPR17, wu2022grit} provides more descriptive language descriptions in a generative way. Scene graph generation~\cite{li2017scene, tang2018learning, yang2022panoptic} analyzes the relationships between regions by the graph. The VCR~\cite{zellers2019vcr} dataset presents many region-level reasoning cases and ~\cite{yu2021ernievil, su2019vlbert, li2019visualbert, yao2022pevl} exhibit decent performance by correctly selecting the answers in the multiple-choice format. However, a general-purpose region understanding model has yet to emerge. In this paper, by harnessing the powerful large language model~\cite{touvron2023llama, vicuna2023},  \ours uses a generative approach to handle all these tasks. Users can complete various region-level understanding tasks by freely asking questions.

\subsection{Other Region Reference Format.}
 Many recent studies (\textbf{concurrent works or follow-ups of \ours}) integrate region-level data via different region reference formats into the training of MLLMs~\cite{peng2023kosmos2, chen2023shikra, zhao2023bubogpt, pi2023perceptiongpt, pramanick2023jack, you2023ferret, yuan2023osprey}. In particular, Kosmos-2~\cite{peng2023kosmos2}, Shikra~\cite{chen2023shikra}, and CogVLM~\cite{wang2023cogvlm} directly quantize bounding boxes into discrete location tokens or numeric representations of positions. Regiongpt~\cite{guo2024regiongpt}, Osprey~\cite{yuan2023osprey}, and Groma~\cite{ma2024groma} use different feature extraction operations to extract the features of user reference, which may be partially influenced by our approach.  Additionally, there has been research that explores a fusion of the two approaches, as shown in references~\cite{you2023ferret, rasheed2023glamm, zhang2023next}.

\section{\ours}

\begin{figure*}[t]
  \centering
  \includegraphics[width=\textwidth]{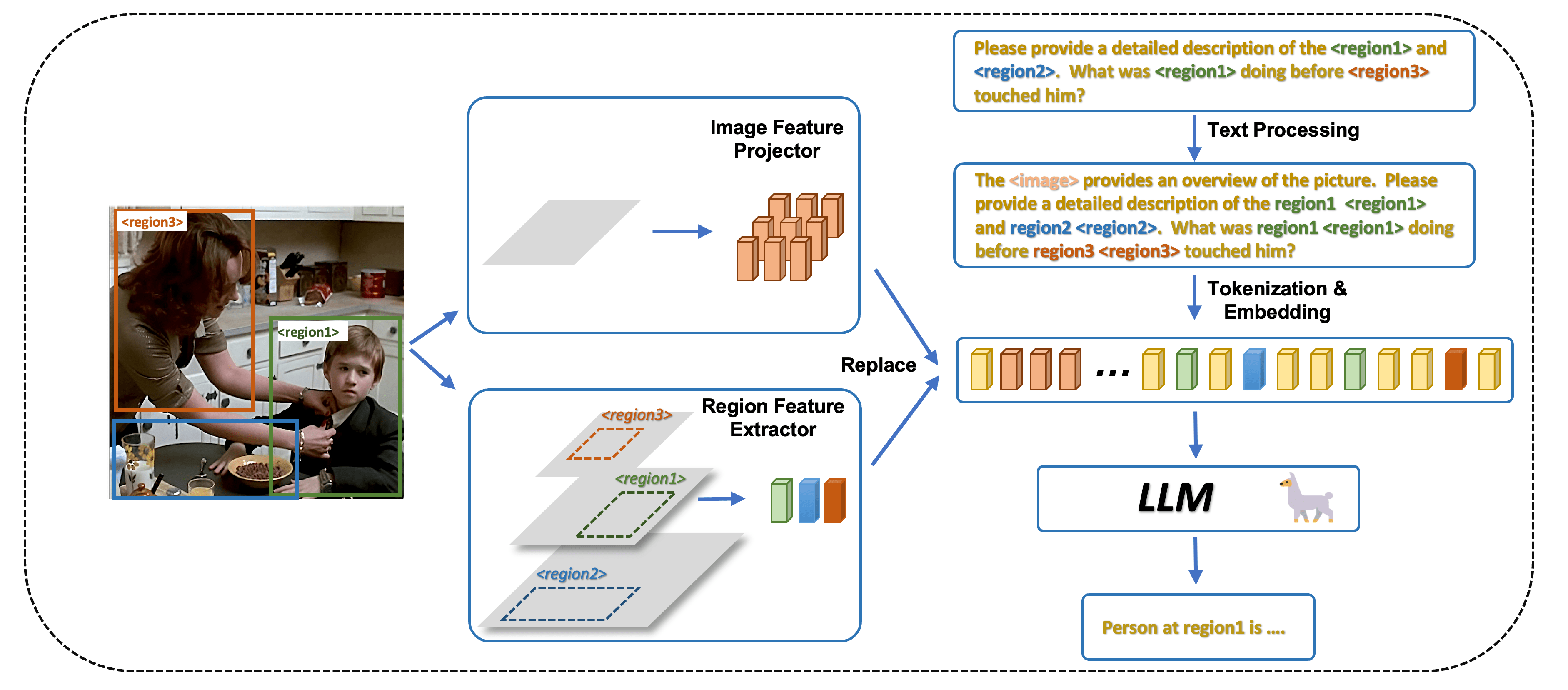}
  \caption{\ours is an end-to-end vision-language model for processing spatial instructions that contain references to the region-of-interest, such as \textit{<region\{i\}>}. During tokenization and conversion to embeddings, the embedding of \textit{<region\{i\}>} in the instruction is replaced with the RoIAlign results from multi-level image features. Subsequently, such an interleaved region feature and language embedding sequence can be sent to a large language model (LLM) for further processing. We also utilize the entire image feature to capture global information.}
  \label{architecture}
\end{figure*}

\ours consists of a vision encoder, a projector for image-level features, a region feature extractor, and a large language model (LLM). Compared to previous works~\cite{zhu2023minigpt, liu2023visual}, \ours stands out for its ability to convert instructions that include spatial positions into an interleaved sequence of region features and text embeddings, as shown in Figure~\ref{architecture}.

\subsection{Model Architecture}
We adopt the CLIP ViT-L/14~\cite{radford2021learning} as the vision encoder. We use the feature map of the penultimate transformer layer as the representation of the entire image and then map it to the language space using a single linear layer. A region feature extractor is used to obtain region features with the region references in the user instruction. Finally, we employ the Vicuna~\cite{zheng2023judging}, an instruction-tuned LLaMA~\cite{touvron2023llama}, to perform further processing.

We utilize widely adopted modules in the field of object detection to construct the region feature extractor. To ensure a robust feature representation for regions of varying scales, we construct a multi-level image feature pyramid~\cite{lin2017feature} by selecting four layers from the CLIP vision encoder and fusing them with five lightweight scale shuffle modules~\cite{Zhang_2023_CVPR}. These layers are located at the second-to-last, fifth-to-last, eighth-to-last, and eleventh-to-last positions, respectively. Additionally, we incorporate feature coordinates~\cite{liu2018intriguing, wang2020solo} for each level to address the problem of translation invariance in CNNs. This helps make the model sensitive to absolute position information, such as the description \textsl{``girl on left''} in Figure~\ref{fig:vis_stage1}. Finally, we use RoIAlign to extract region-level features with an output size of 14$\times$14~\cite{he2017mask}, which ensures that sufficient detailed information is preserved. Moreover, all four level features are involved in the RoIAlign operation and fused into a single embedding as the representation of the region of interest (RoI)~\cite{liu2018path}.

\subsection{Tokenization and Embedding} 
To enable users to refer to regions of interest in text inputs, we define a special token \textsl{<region\{i\}>}, which acts as the placeholder that will be replaced by the corresponding region feature after tokenization and embedding. One example is depicted in Figure~\ref{architecture}. When a user presents a spatial instruction, \textsl{``What was \texttt{<region1>} doing before \texttt{<region3>} touched him?''}, the embedding of \texttt{<region1>} and \texttt{<region3>} are replaced by their corresponding region features. However, this replacement discards the pure text references to different regions. To allows LLM to maintain the original references~(\textsl{region1, region3}) in the response sequence, the instruction is modified to \textsl{``What was region1 \texttt{<region1>} doing before region3 \texttt{<region3>} touched him?''}. Then, LLM can generate a reply like \textsl{``The person in region1 was eating breakfast before the person in region3 touched them.''}

Regardless of the user instruction, we incorporate a prefix prompt, \textsl{``The \texttt{<image>} provides an overview of the picture.''} The \textsl{\texttt{<image>}} is a special token that acts as a placeholder, the embedding of which would be replaced by image features of the vision encoder. These features enable LLM to receive comprehensive image information and obtain a holistic understanding of the visual context.
\subsection{Spatial Instruction Tuning}
\label{sec:data}

Our model is trained using a next-token prediction loss~\cite{liu2023visual, zhu2023minigpt}, where the model predicts the next token in a given input text sequence.

We formulate annotations in instruction tuning format by creating a question that refers to the mentioned region for each region-text annotation. We partition the available region-text data into two groups and use them separately in two distinct training stages. In the first stage, we attempt to align region feature with word embedding in language models using simple region-text pairs that contain color, position, or category description. The second stage is designed to handle more complex concepts, such as action, relationship, and common sense reasoning. Furthermore, we provide diverse instructions for these datasets to simulate chat-like input in this stage. 

\noindent \textbf{Stage 1: Pre-training} In this stage, we first load the weights of LLaVA~\cite{liu2023visual} after its initial stage of training, which includes a pre-trained vision encoder, a projector for image-level features, and an LLM. We only keep the region feature extractor trainable and aim to align region features with language embedding by collecting short text and bounding box pairs. These pairs are from both normal detection datasets and referring expression detection datasets, which have short expressions. The objective is to enable the model to recognize categories and simple attributes of the region in an image, which are typically represented by a short text annotation~(usually within 5 words). Specifically, we utilize  COCO~\cite{lin2014microsoft}, RefCOCO~\cite{yu2016modeling}, and RefCOCO+~\cite{yu2016modeling} datasets in this stage.

As shown in Table~\ref{tab:stage1}, for COCO detection data, we first explain the task in the prompt and then convert the annotations to a single-word region caption task. For RefCOCO and RefCOCO+, we also give task definitions first and train the model to generate descriptions containing basic attributes of the region. Only the description of the region (\textcolor[rgb]{0.8,0,0}{in red color}) will be used to calculate the loss.

\begin{table}[t]
\centering
\caption{Exampls for Stage 1 training data: For both tasks, we begin by providing a description of the task and the expected answer. Then, we concatenate all region-text pairs into a sequence. For detection data, the format is \textsl{<region\{i\}> category\_name}. For referring expression comprehension, the format is \textsl{<region\{i\}> description of region}. Only the responses highlighted  \textcolor[rgb]{0.8,0,0}{in red} are used to calculate the loss.}
\resizebox{\linewidth}{!}{
\begin{tcolorbox}[colback=white!100] {
\centering
\footnotesize
{\begin{tabular}{p{0.97\columnwidth} c}

{\bf Object Detection:} \\ 
In the conversation below,  you simply answer the category name based on what you see in the imagery inside a particular region. I \\ 
 will  give you only one region \\
each time. Categories \\
containing person, bicycle, \\ car  ... \\
<region1> \textcolor[rgb]{0.8,0,0}{person},  \\
<region2> \textcolor[rgb]{0.8,0,0}{dog} \\
\multirow{-6.5}{*}
{\hspace{6cm}\includegraphics[height=2.2cm]{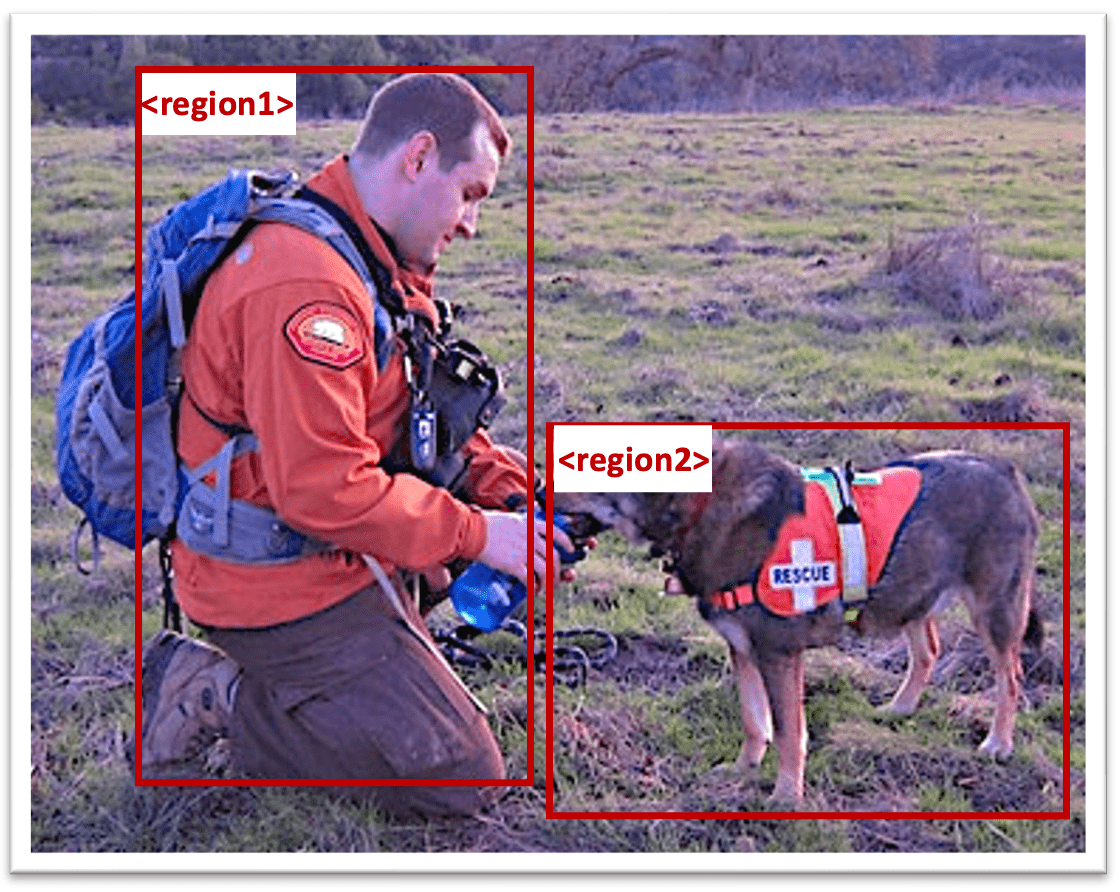}}

\vspace{-0.3cm} 
\hrulefill & \\

{\bf Referring Expression Comprehension} &\\
I will provide you with only one region containing only one object, although there may be other objects present in the image. It is 
recommended that you describe the object's relative \\
position with respect to  other\\
objects in the image and its \\
basic attributes.\\
<region1> \textcolor[rgb]{0.8,0,0}{{red shirt girl}} \\
<region2> \textcolor[rgb]{0.8,0,0}{guy in black} \\
<region3> \textcolor[rgb]{0.8,0,0}{right  most person } \\
\multirow{-7.5}{*}
{\hspace{6cm}\includegraphics[height=2.2cm]{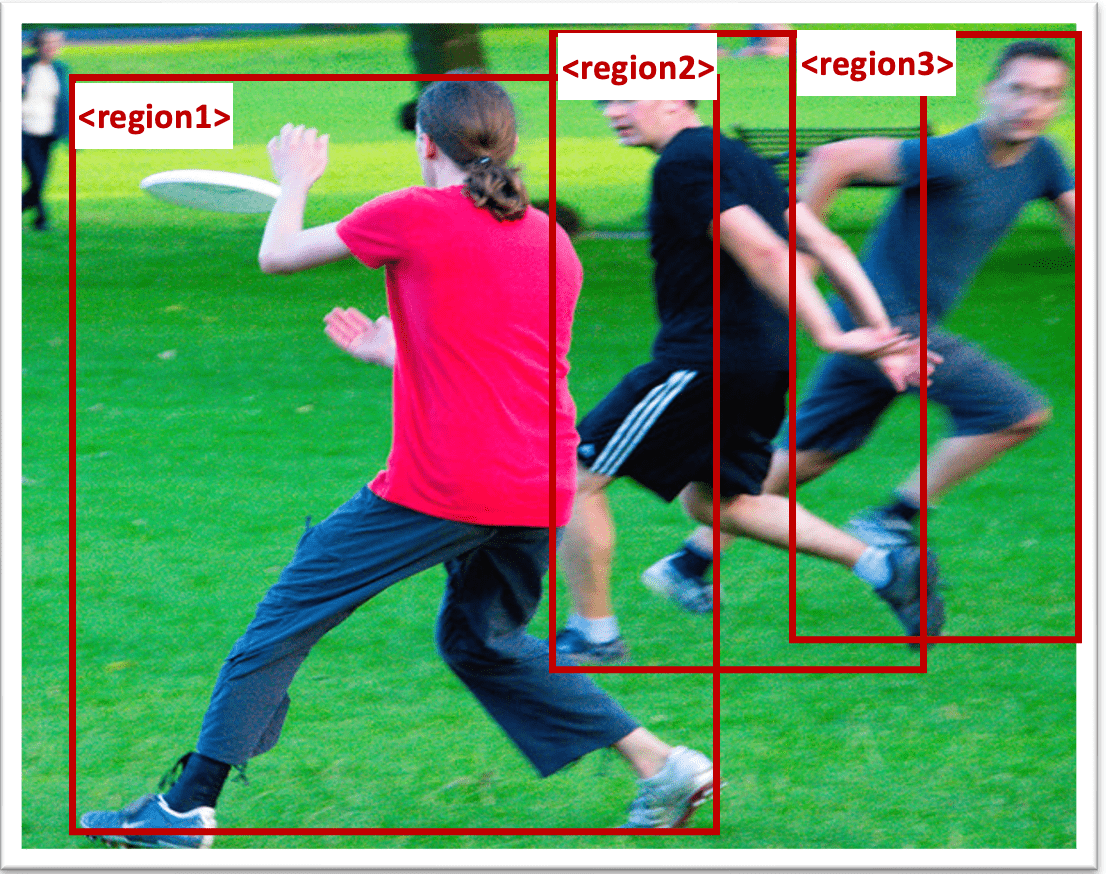}}
\vspace{-6mm}
\end{tabular}}}
\end{tcolorbox} }

\label{tab:stage1}
\end{table}

\begin{figure}[!h]
  \centering
  \includegraphics[width=\textwidth]{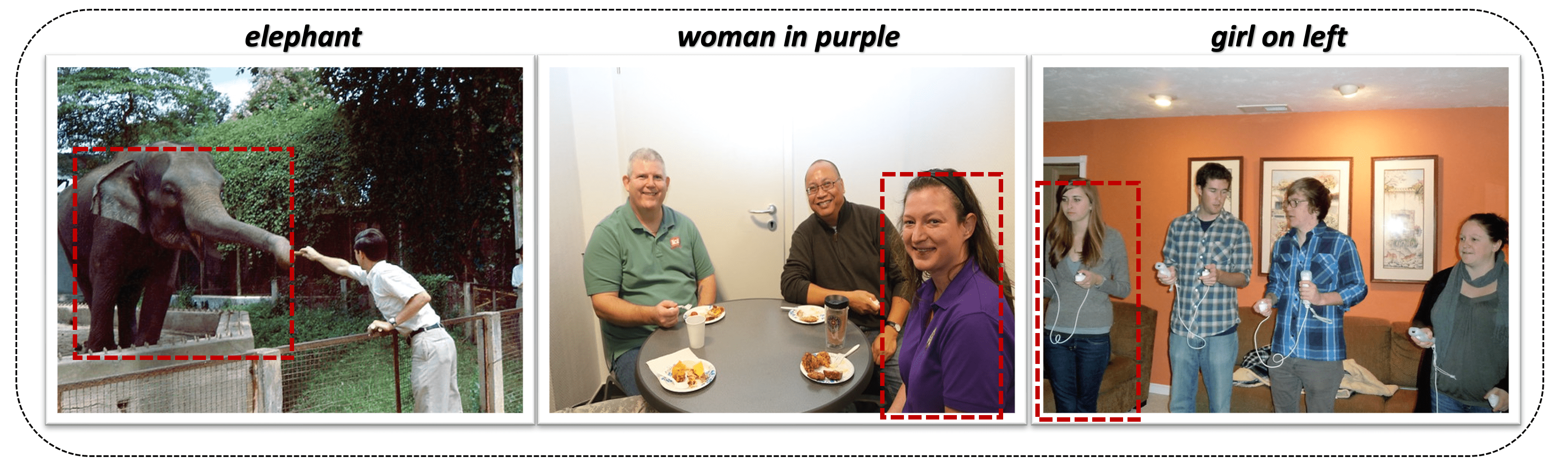}
  \caption{After Stage 1 training, \ours is capable of identifying the category of the region (elephant), simple attributes such as color (purple), and the position of the region (left).}
  \label{fig:vis_stage1}
  \vspace{-4mm}
\end{figure}

After this training stage, \ours can recognize categories, simple attributes, and positions of regions in images, as shown in Figure~\ref{fig:vis_stage1}.

\noindent \textbf{Stage 2: End-to-end fine-tuning} In this stage, we only keep the vision encoder weights fixed and train the region feature extractor, image feature projector, and LLM weights. Our main focus is to enhance \ours's ability to accurately follow user instructions and tackle complex single/multiple region understanding tasks.

\begin{table}[t]
\centering
\caption{Exampls for Stage 2 training data: Only the response \textcolor[rgb]{0.8,0,0}{in red color} and stop string \textbf{\#\#\#} will be used to calculate the loss.}
\resizebox{\linewidth}{!}{
\begin{tcolorbox}[colback=white!100]
\centering
\footnotesize
\begin{tabular}{p{0.97\columnwidth} c}
{\bf  Region Caption } &\\
\#\#\# Question: Can you provide me with a detailed description of the region &\\
marked by <region1> ?\\
\#\#\# Answer: \textcolor[rgb]{0.8,0,0}{{A man wearing}} \\
 \textcolor[rgb]{0.8,0,0}{{a light blue T-shirt and jeans}} \\
\textcolor[rgb]{0.8,0,0}{{ with his arms extend. }} \\
\hspace{6cm} 
\multirow{-5}{*}
{\includegraphics[height=2.2cm]{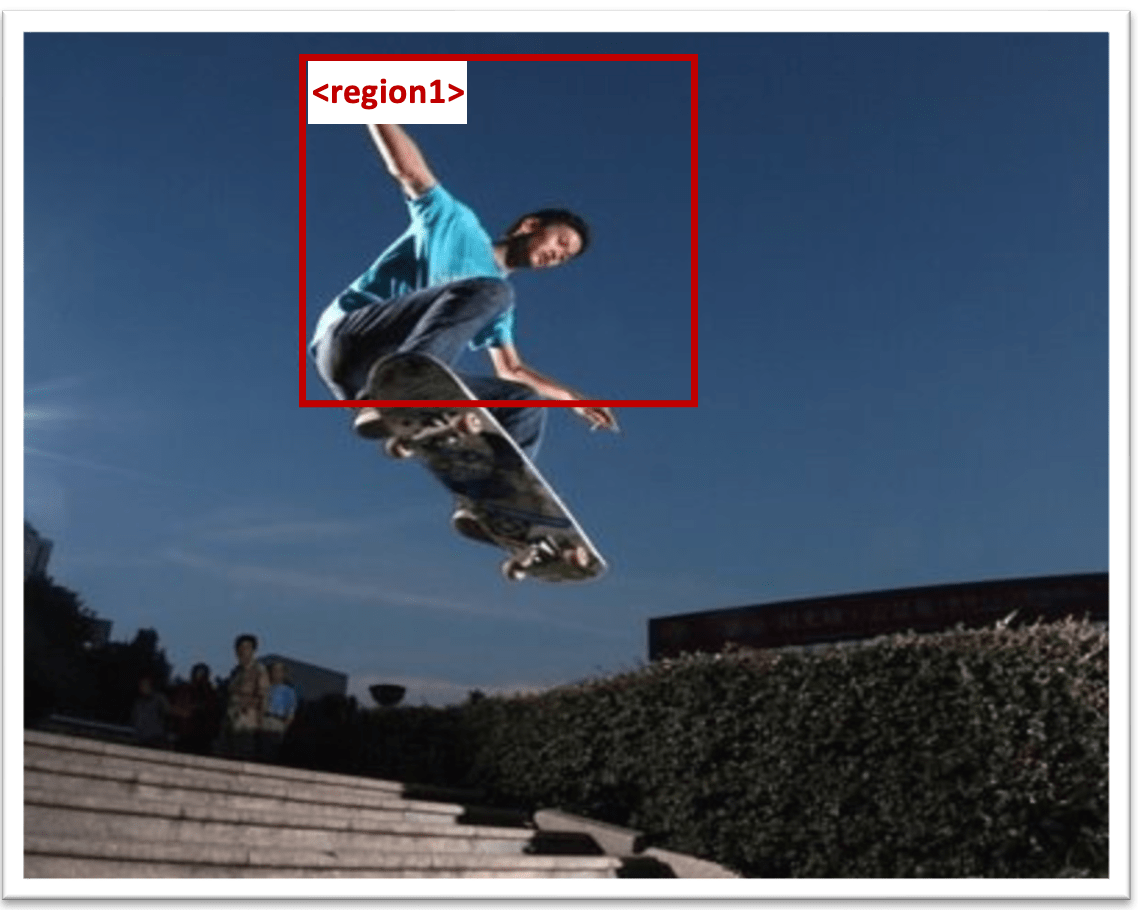}}

\vspace{3mm}
\hrulefill & \\

\#\#\# Question: Could you please give me a detailed description of areas <region1>, <region2>, <region3>, <region4>, \\ <region5> ?\\
\#\#\# Answer: \textcolor[rgb]{0.8,0,0}{{A man in a white }} \\
\textcolor[rgb]{0.8,0,0}{{ shirt with a plate of food sits }} \\
\textcolor[rgb]{0.8,0,0}{{ outside in a folding chair with }} \\
\textcolor[rgb]{0.8,0,0}{{ a little girl who is writing}} \\
\hspace{6cm} 
\multirow{-5}{*}
{\includegraphics[height=2.2cm]{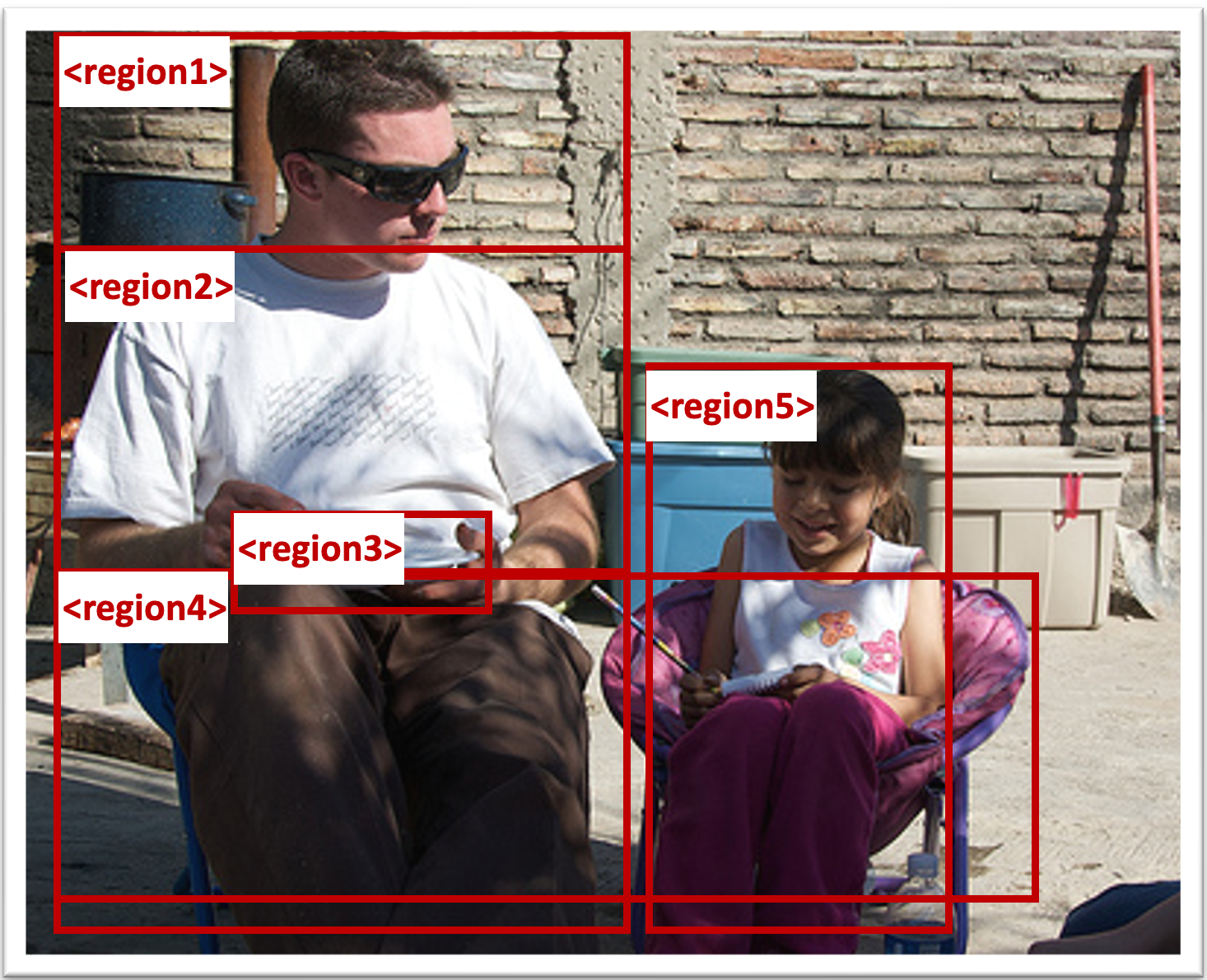}}

\vspace{3mm}
\hrulefill & \\

{\bf  Region Reasoning} &\\

\#\#\# Question: Is <region1> happy to be speaking with <region2> and <region3> ? \\
\#\#\# Answer: \textcolor[rgb]{0.8,0,0}{{No, person at region1 is bothered by the conversation.}} \\

\#\#\# Question: What factors \\
influenced your perspective? \\
\#\#\# Answer: \textcolor[rgb]{0.8,0,0}{{Person at region1 }} \\
\textcolor[rgb]{0.8,0,0}{is standing with his hand on } \\
\textcolor[rgb]{0.8,0,0}{his hip in a defensive way.} \\
\hspace{6cm} 
\multirow{-6}{*}
{\includegraphics[height=2.2cm]{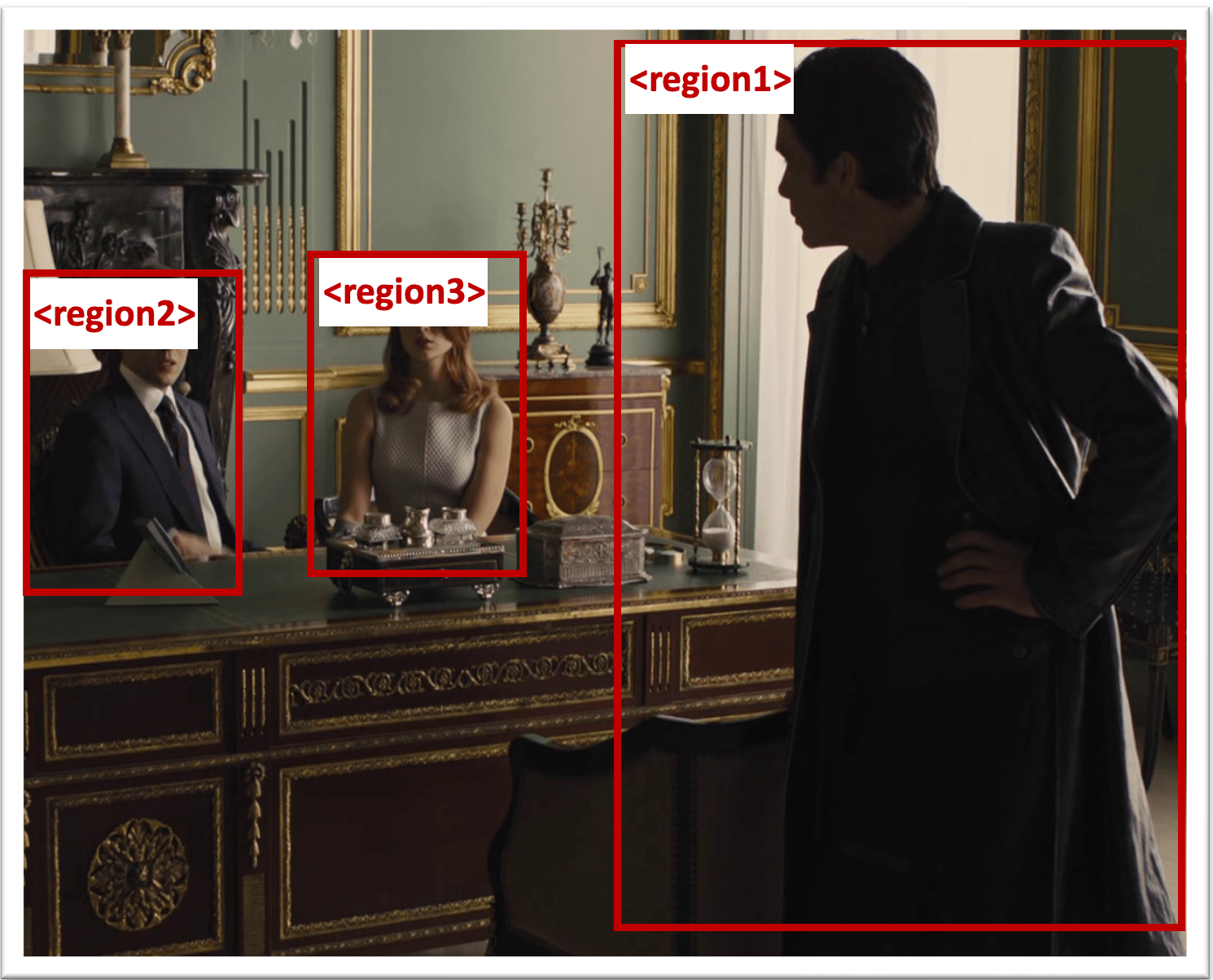}}
\vspace{-5mm}

\end{tabular}
\end{tcolorbox}}

\label{tab:stage2}
\end{table}

As shown in Table~\ref{tab:stage2}, we tailor specific instructions for different tasks. For single region caption, we construct from Visual Genome~(VG) region caption part~\cite{krishna2017visual} and RefCOCOg~\cite{mao2016generation}. For multiple region caption, Flicker30k~\cite{plummer2015flickr30k} is converted to a multiple region caption task where the caption should include all visual elements emphasized by bounding boxes. To simulate user instruction, we create 20 questions for each caption task. For the region reasoning task, we modify Visual Commonsense Reasoning (VCR)~\cite{zellers2019recognition} to meet the input format requirements and make it more similar to human input. 

\begin{figure*}[!t]
  \centering
  \includegraphics[width=1.\textwidth]{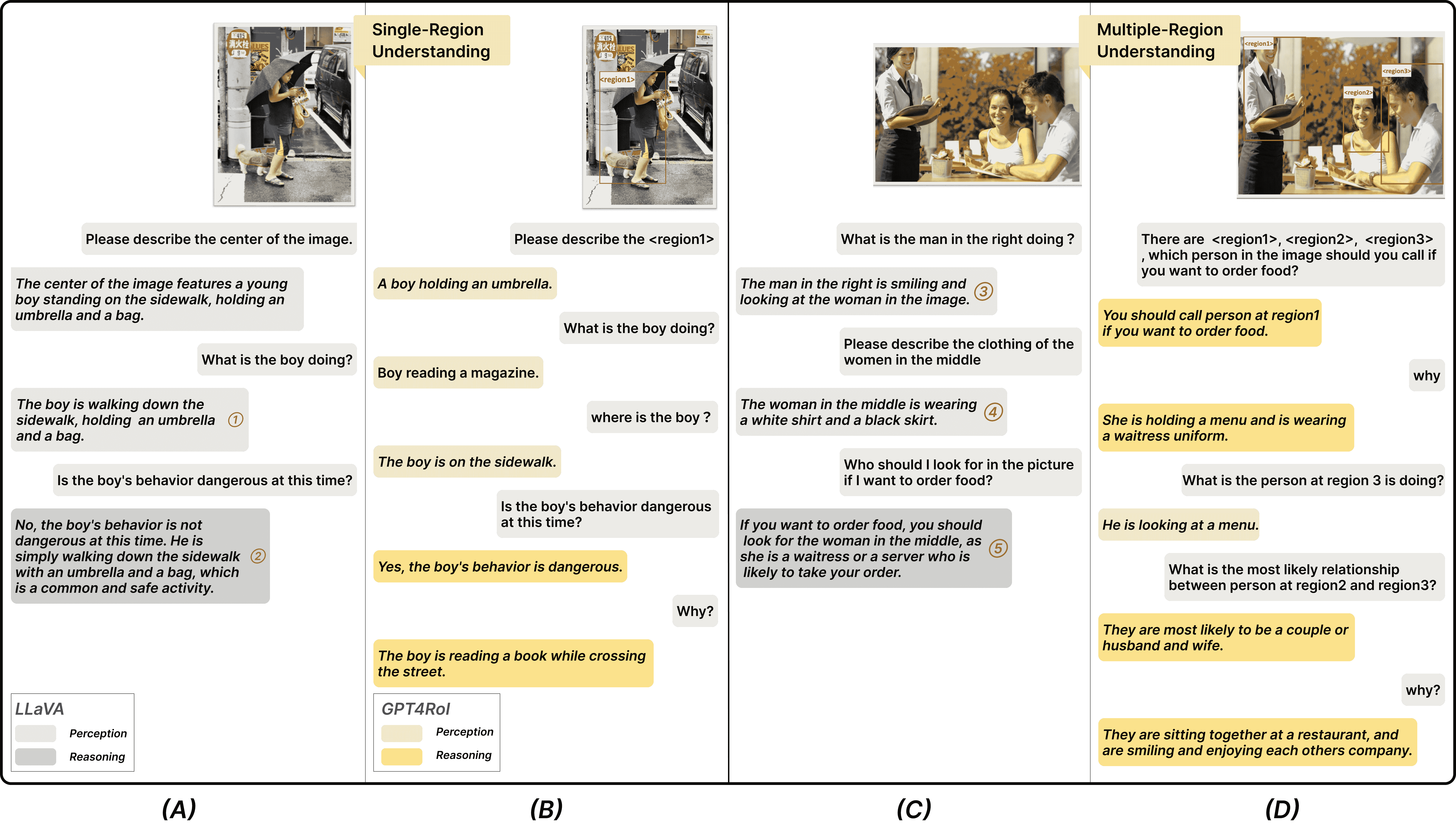}
  \caption{\ours and LLaVA dialogue performance showcase. Figures A and C demonstrate the dialogue scenarios of LLaVA when referring to a single instance and multiple instances solely using natural language in the conversation. On the other hand, Figures B and D showcase how GPT4RoI utilizes bounding boxes as references to address the same scenarios. }
  \label{fig:demo}
\vspace{-3mm}
\end{figure*}

To improve the capability of \ours for multi-round conversation and generate more human-like responses, we also involve the LLaVA150k~\cite{liu2023visual} visual instruction dataset in this stage. We employ an off-the-shelf LVIS detector~\cite{EVA02} to extract up to 100 detection boxes per image. These boxes are then concatenated with the user instructions in the format \textsl{``<region\{i\}> may feature a class\_name''}. LLaVA150k significantly improves the capability of \ours for multi-round conversation.

After completing this training stage, \ours is capable of performing complex region understanding tasks based on user instructions, including region caption and reasoning, as demonstrated in Figure~\ref{fig:demo} and Section~\ref{demos}.

\section{Demostrations}
\label{demos}

In this section, we compare the differences between the \emph{visual} instruction tuning model LLaVA~\cite{liu2023visual} and our \emph{spatial} instruction tuning model GPT4RoI. We demonstrate our new interactive approach and highlight its advanced capabilities in understanding multimodality.

As shown in Figure~\ref{fig:demo}.A, when we try to make LLaVA focus on the center region of the image, it only sees the boy holding an umbrella and a bag, but it misses the book. As a result, LLaVA gives a wrong answer to the question \emph{``What is the boy doing''} (Figure~\ref{fig:demo}.A.\ding{172}), and this leads to an incorrect conclusion that \emph{``the boy's behavior is not dangerous''} (Figure~\ref{fig:demo}.A.\ding{173}). 

In comparison, as shown in Figure~\ref{fig:demo}.B, our approach \ours efficiently recognizes visual details using the given bounding box. This allows it to accurately identify the action of  \emph{``reading a magazine.''} Furthermore, \ours demonstrates its reasoning abilities by correctly inferring that the  \emph{``boy's behavior is dangerous''}, and giving a reasonable reason that \emph{``the boy is reading a book while crossing the street''}.

When there are multiple instances in the image (as depicted in Figure~\ref{fig:demo}.C), we attempt to refer to the corresponding instances as \emph{``the right''} and \emph{``the middle''}. However, LLaVA provides incorrect information by stating that the right man is \emph{``looking at the women''} (as shown in Figure~\ref{fig:demo}.C.\ding{174}). Even more concerning, LLaVA overlooks the actual women in the middle and mistakenly associates the women on the left as the reference, resulting in completely inaccurate information (as shown in Figure~\ref{fig:demo}.C.\ding{175} \& \ding{176}).

In comparison, as shown in Figure~\ref{fig:demo}.D, \ours is able to understand the user's requirements, such as identifying the person to call when ordering food, and accurately recognize that the person in region1 fulfills this criterion. Additionally, it correctly recognizes that the person in region3 is \emph{``looking at the menu''}. Importantly, \ours can also infer relationships between the provided regions based on visual observations. For example, it deduces that the likely relationship between region2 and region3 is that of a \emph{``couple''}, providing a reasonable explanation that they \emph{``are smiling and enjoying each other's company''}.

\section{Experiments}
 We adopt several representative benchmarks to quantitatively evaluate the region understanding ability of GPT4RoI. First, we give the results on comprehensive LLM image region understanding benchmark ViP-Bench~\cite{cai2023making}. We also evaluate the region recognition ability through the open-vocabulary region recognition and region caption task. For region reasoning ability, we report the results on  Visual Commonsense Reasoning~\cite{zellers2019recognition}. 

\subsection{Comprehensive Region Understanding}
ViP-Bench~\cite{cai2023making} is a comprehensive benchmark for evaluating multimodal models' region understanding capabilities, including recognition~(Rec), OCR, knowledge~(Know), math, relationship~(Rel), and language generation~(Lang). As shown in Table~\ref{tab:vipbench}, we give the comparison of methods including InstructBLIP~\cite{instructblip} with a pure language reference, Shikra ~\cite{chen2023shikra} with textual coordinates as reference, Kosmos-2 ~\cite{peng2023kosmos2} with extra position tokens. We can see that GPT4RoI surpasses by a clear margin with actually much less training data.

\begin{table}[t]
\centering
\caption{Evaluation results on ViP-Bench.The assessed dimensions are Recognition (Rec), OCR, Knowledge (Know), Math, Relationship (Rel), and Language Generation (Lang). $^\dagger$ means concurrent work.}
\footnotesize
{
 \begin{tabular}
{l@{\hspace{20pt}}  p{10mm}p{10mm}p{10mm}p{10mm}p{10mm}p{10mm}p{10mm}p{10mm}p{10mm}}
\toprule
\multirow{1}{*}{Type}  & Rec & OCR & Know  & Math & Rel & Lang & All \\ 
\midrule
 Kosmos-2$^\dagger$   & 29.5    & 14.2  & 18.5	 & 9.7  & 7.5	 & 21.9 & 26.9   \\
 InstructBLIP     & 36.9    & 16.3  & 34.2    & 22.3 & 26.8   & 7.5 & 31.7  \\
 Shikra-7B$^\dagger$    & 40.2	   & 10.0  & 28.0    & 3.5  & 18.9	 & 20.6 & 33.7  \\
\midrule
GPT4RoI-7B 	&35.6 &  16.7	 & 29.7	 & 9.7 & 	32.5 & 	13.8 & 35.1    \\
\bottomrule
\end{tabular}
}

\label{tab:vipbench}
\end{table}

\subsection{Region Recognization}
We also conduct evaluations on several specific region-level understanding tasks to demonstrate the superior performance of \ours. The instruction template for these tasks is provided in Table~\ref{tab:ds_prompt}.
\begin{table}[!h]
\centering
\caption{Task prompt of three downstream tasks.}
\resizebox{0.8\linewidth}{!}{
\begin{tcolorbox}[colback=white!100]
\centering
\footnotesize
\begin{tabular}{p{1\columnwidth} c}
\textbf{Region Caption Task } \\ \\
\#\#\# Question: Can you give a description of the region mentioned by <region> \\ \\
\#\#\# Answer: \textcolor[rgb]{0.8,0,0}{{A man wearing a light blue t-shirt and jeans with his arms extended}} \\

\\
\textbf{Region Reasoning Task on VCR} \\ \\
\textbf{Q $\rightarrow$ A} \\

\#\#\# Question: <region1>,<region2>,<region3>... refers to specific areas within the photo along with their respective identifiers. I need you to answer the question. Questions are multiple-choice; you only need to pick the correct answer from the given options (A), (B), (C), or (D). \\ \\
How is 1 feeling ? \\
(A),1 is feeling amused . \\
(B),1 is upset and disgusted . \\
(C),1 is feeling very scared . \\
(D),1 is feeling uncomfortable with 3 \\ \\
\#\#\# Answer: \textcolor[rgb]{0.8,0,0}{{(C)}} \\ \\

\textbf{QA $\rightarrow$ R} \\
\#\#\# Question: <region1>,<region2>,<region3>... refers to specific areas within the photo along with their respective identifiers. 
I give you a question and its answer, I need you to provide a rationale explaining why the answer is right. Both questions are multiple-choice; you only need to pick the correct answer from the given options (A), (B), (C), or (D). \\ \\
"How is 1 feeling ?" The answer is "1 is feeling very scared." What's the rationale for this decision? \\
(A),1's face has wide eyes and an open mouth . \\
(B),When people have their mouth back like that and their eyebrows lowered they are usually disgusted by what they see . \\
(C),3,2,1 are seated at a dining table where food would be served to them . people unaccustomed to odd or foreign dishes may make  disgusted looks at the thought of eating it . \\
(D),1's expression is twisted in disgust . \\ \\

\#\#\# Answer: \textcolor[rgb]{0.8,0,0}{{(A)}} \\
\end{tabular}
\end{tcolorbox}}

\label{tab:ds_prompt}
\end{table}

\noindent\textbf{Open-Vocabulary Recognition} The results are obtained by calculating the semantic similarity between the generated region caption of GPT4RoI and the vocabulary lists of each dataset and then selecting the category with the highest similarity as the final result~(as the study~\cite{yuan2023osprey}). The input to the CLIP-Surgery-ViT-L is the cropped region with a size of 512$\times$512. As shown in Table~\ref{tab:ovr}, constrained by the resolution of 224$\times$224, GPT4RoI exhibits slightly lower performance than CLIP-Surgery-ViT-L only at the instance level. Considering all metrics score on two datasets, GPT4RoI demonstrates robust and comprehensive region recognition capabilities compared to our concurrent works that employ alternative reference formats. 

\begin{table}[t]
  \centering
    \caption{Recognition performance on panoptic segmentation (PQ), instance segmentation (AP) and semantic segmentation (mIoU) upon the validation sets of Cityscapes~\cite{cordts2016cityscapes} and ADE20K~\cite{zhou2017scene}. $^\dagger$ means concurrent work.}
{
  \begin{tabular}
  {l@{\hspace{20pt}}  p{10mm}p{10mm}p{10mm}p{10mm}p{10mm}p{10mm}p{10mm}p{10mm}p{10mm}}
    \toprule
    \multirow{2}{*}{\textbf{Method}} & \multicolumn{3}{c}{Cityscapes}&\multicolumn{3}{c}{ADE20K-150} \\
    \cmidrule(lr){2-4} \cmidrule(lr){5-7}  & PQ & AP & mIoU & PQ & AP & mIoU \\
    \midrule
    CLIP-Surgery-ViT-L   & 27.24  & 28.35& 21.92 & 26.55 & 29.70 & 21.42 \\
    Kosmos-2$^\dagger$   & 12.09 & 9.81 & 13.71 & 6.53& 4.33 & 5.40  \\
    Shikra-7B$^\dagger$    & 17.80 & 11.53  & 17.77 & 27.52 & 20.35 & 18.24 \\
    \midrule
    GPT4RoI-7B   & 34.70 & 21.93 & 36.73 & 36.32 & 26.08 & 25.82  \\

    \bottomrule
  \end{tabular}
  }

  \label{tab:ovr}
\end{table}

\begin{table}[t]
\centering
\caption{Compariation of region caption ability on the validation dataset on Visual Genome. All methods employ ground truth bounding boxes as input. $^\dagger$ means concurrent work. $^\diamond$ means the model is after fine-tuning on Visual Genome.}\label{tab:region_cap}
{
\begin{tabular}
  {l@{\hspace{20pt}}  p{15mm}p{15mm}p{15mm}p{15mm}}
\toprule
\multicolumn{1}{c}{Model}  & \multicolumn{1}{c}{BLEU@4} & \multicolumn{1}{c}{METEOR}& \multicolumn{1}{c}{ROUGE}& \multicolumn{1}{c}{CIDEr} \\ 

\midrule
GRiT & - & 17.1 & -  & 142.0 \\
Shikra-7B$^\dagger$ & 8.9  & 15.2 & 30.3  & 115.8\\
\midrule
GPT4RoI-7B & 10.5  & 16.5 & 33.4  & 134.5\\
GPT4RoI-7B$^\diamond$ & 11.5  & 17.4 & 35.0  & 145.2\\
GPT4RoI-13B$^\diamond$ & 11.7  & 17.6 & 35.2 & 146.8 \\

\bottomrule
\end{tabular}
}

\end{table}

\begin{table}[t]
\centering
\caption{Accuracy scores on VCR. \ours achieves state-of-the-art accuracy among all methods.}\label{tab:vcr}

    \small
\resizebox{\linewidth}{!}{\begin{tabular}{lcccccccc}  
    \toprule
    \multirow{2}{*}{Model} & \multirow{2}{*}{\rebuttal{Code}}  &\multirow{2}{*}{\rebuttal{\#Params}}  & \multicolumn{3}{c}{Val Acc.(\%)} & \multicolumn{3}{c}{Test Acc.(\%)} \\ 
    \cmidrule(lr){4-9}
     &  & & Q $\rightarrow$ A & QA $\rightarrow$ R & Q $\rightarrow$ AR & Q $\rightarrow$ A& QA $\rightarrow$ R & Q $\rightarrow$ AR \\
    \midrule
    ViLBERT~\cite{lu2019vilbert} &  \rebuttal{Y} & \rebuttal{221M} & 72.4  & 74.5 &54.0  &73.3 & 74.6 & 54.8 \\
    Unicoder-VL~\cite{li2019unicodervl} & \rebuttal{ Y} & - & 72.6  & 74.5& 54.5 &73.4 & 74.4 & 54.9 \\
    VLBERT-L~\cite{su2019vlbert}   &  \rebuttal{ Y} & \rebuttal{383M}  & 75.5 & 77.9 & 58.9 &75.8 & 78.4 & 59.7 \\
    UNITER-L~\cite{chen2020uniter} &  \rebuttal{ Y} & \rebuttal{303M}  & -&-&- &77.3&80.8&62.8\\
    ERNIE-ViL-L~\cite{yu2021ernievil} &  \rebuttal{ Y} & - & 78.52 & 83.37 & 65.81 &79.2&83.5&66.3 \\
    MERLOT~\cite{zellersluhessel2021merlot} &  \rebuttal{ Y} &  \rebuttal{223M} &- &-&-&80.6&80.4&65.1 \\
    VILLA-L~\cite{gan2020largescale} &  \rebuttal{ Y} & - &78.45 & 82.57 & 65.18 & 78.9& 82.8 & 65.7 \\
    RESERVE-L~\cite{zellers2022merlotreserve} &  \rebuttal{ Y} & \rebuttal{644M} & - &-&-&84.0&84.9&72.0\\
    VQA-GNN-L~\cite{wang2022vqagnn} &  \rebuttal{ Y} & \rebuttal{ 1B+} & - &-&-&85.2&86.6& 74.0\\
    \midrule
    GPT4RoI-7B~(ours)  &  \rebuttal{ Y} &  \rebuttal{7B+} &\textbf{87.4} & \textbf{89.6} & \textbf{78.6}  & -  & - & -  \\
    \midrule
    VLUA+@Kuaishou &  \rebuttal{ N} & - & - &-&-& 84.8 & 87.0 & 74.0 \\
     KS-MGSR@KDDI Research and SNAP & \rebuttal{ N} & - &-&-&-& 85.3&86.9 & 74.3 \\
    SP-VCR@Shopee & \rebuttal{ N} & - & - &-&-& 83.6 & 88.6 & 74.4\\
    HunYuan-VCR@Tencent & \rebuttal{ N} & - & - & - &- & 85.8 & 88.0 & 75.6 \\
    Human Performance~\cite{zellers2019recognition} & - & - & - & - & -  & 91.0 & 93.0 & 85.0 \\
    \midrule
    GPT4RoI-13B~(ours) & \rebuttal{Y} & \rebuttal{ 13B+} & - & - & -  & \textbf{89.4} & \textbf{91.0} & \textbf{81.6} \\
    \bottomrule
    \end{tabular}}

\end{table}

\noindent \textbf{Region Caption}  We report the scores of BLEU, METEOR, ROUGE, and CIDEr for different methods on the validation set of Visual Genome~\cite{krishna2017visual}. 
As shown in Table~\ref{tab:region_cap}, compared to the concurrent work Shikra~\cite{chen2023shikra} with textual coordinates as region reference, \ours shows clearly better region caption ability. After fine-tuning, \ours outperforms the previous state-of-the-art specialist model GRiT~\cite{wu2022grit} by a significant margin, without any additional techniques or tricks. Additionally, we observe that the performance of \ours-7B and \ours-13B is comparable, suggesting that the bottleneck in performance lies in the design of the visual module and the availability of region-text pair data. These areas can be explored further in future work.

\subsection{Region Reasoning}

 Visual Commonsense Reasoning~\cite{zellers2019recognition} employs multiple-choice settings that require both recognition ability and commonsense reasoning to select the correct choice. Calculating the similarity of LLM's output with the correct choice is quite a challenge. This is because it is hard to find a model that can capture the reasoning similarity. Therefore, we finetune GPT4RoI to align with the answer format, following conventional methods.

\noindent\textbf{Visual Commonsense Reasoning} Visual Commonsense Reasoning (VCR) offers a highly demanding scenario that necessitates advanced reasoning abilities, heavily relying on common sense. Given the question(Q), the model's task is not only to select the correct answer(A) but also to select a rationale(R) that explains why the chosen answer is true.  In the Q$\rightarrow$A setup, a model is given a question and must select the correct answer from four choices. In the QA->R setup, a model is provided with a question and the correct answer, and it needs to justify the answer by selecting the most appropriate rationale from four choices. The performance of models is evaluated using the Q$\rightarrow$AR metric, where accuracy is measured as the percentage of correctly answered questions along with the correct rationale.

As shown in Table~\ref{tab:vcr}, \ours shows significant improvements over the previous methods across all $Q\rightarrow A$, $QA\rightarrow R$, and $Q\rightarrow AR$ tasks. Notably, in the crucial $Q\rightarrow AR$ task, \ours-13B achieves a performance of 81.6\% accuracy, surpassing previous methods by over 6 points, even outperforming \rebuttal{confidential} commercial product.  More importantly, this performance is almost reaching human-level performance of 85.0\% accuracy, which shows that the multimodal ability of \ours is promising to be further developed to human intelligence. \rebuttal{Furthermore, comparing GPT4RoI to previous methods, particularly observing the size of the language model used, also demonstrates the significant benefits of the Large Language Model~(LLM) for visual reasoning tasks.}

\section{Conclusions}
 We present \ours, an end-to-end vision-language model that can execute user instructions to achieve region-level image understanding. Our approach employs spatial instruction tuning for the large language model (LLM), where we convert the reference to bounding boxes from user instructions into region features. These region features, along with language embeddings, are combined to create an input sequence for the large language model. We show that \ours enhances user interaction by accurately referring to regions and achieves impressive performance in region-level image understanding tasks.
\section{Acknowledgement}
This project is supported by the National Key R\&D Program of China \\ No.2022ZD0161000.

\clearpage  

%
%
\bibliographystyle{splncs04}
\bibliography{main}
\end{document}